\documentclass[oribibl]{llncs}
\usepackage[utf8]{inputenc}
\usepackage{graphicx}
\usepackage{todonotes}
\usepackage{amsmath}
\usepackage{algorithm}
\usepackage{algpseudocode}
\usepackage{hyperref}
\usepackage{listings}
\usepackage{chngcntr}
\usepackage{subfig}

\title{Densifying Assumed-sparse Tensors\thanks{Accepted to the 2019 International Supercomputing Conference (ISC)}}
\subtitle{Improving Memory Efficiency and MPI Collective Performance during Tensor Accumulation for Parallelized Training of Neural Machine Translation Models}
\author{Derya~Cavdar\inst{3} \and Valeriu~Codreanu\inst{4} \and Can~Karakus\inst{3} \and John~A.~Lockman~III\inst{1} \and Damian~Podareanu\inst{4} \and Vikram~Saletore\inst{5} \and Alexander~Sergeev\inst{2} \and Don~D.~Smith~II\inst{1} \and Victor~Suthichai\inst{3} \and Quy~Ta\inst{1} \and Srinivas~Varadharajan\inst{1} \and Lucas~A.~Wilson\inst{1} \and  Rengan~Xu\inst{1} \and Pei~Yang\inst{1}}
\institute{Dell EMC, Austin TX, USA \and Uber, Seattle WA, USA \and Amazon, Seattle WA, USA \and SURFSara, Utrecht, NL \and Intel, Portland OR, USA}

\begin{document}
\counterwithout{lstlisting}{chapter}

\maketitle

\begin{abstract}
Neural machine translation - using neural networks to translate human language - is an area of active research exploring new neuron types and network topologies with the goal of dramatically improving machine translation performance. Current state-of-the-art approaches, such as the multi-head attention-based transformer, require very large translation corpuses and many epochs to produce models of reasonable quality. Recent attempts to parallelize the official TensorFlow ``Transformer'' model across multiple nodes have hit roadblocks due to excessive memory use and resulting out of memory errors when performing MPI collectives.

This paper describes modifications made to the Horovod MPI-based distributed training framework to reduce memory usage for transformer models by converting assumed-sparse tensors to dense tensors, and subsequently replacing sparse gradient gather with dense gradient reduction. The result is a dramatic increase in scale-out capability, with CPU-only scaling tests achieving 91\% weak scaling efficiency up to 1200 MPI processes (300 nodes), and up to 65\% strong scaling efficiency up to 400 MPI processes (200 nodes) using the Stampede2 supercomputer.
\end{abstract}

\section{Introduction}
Neural Machine Translation (NMT) \cite{2014.Bahdanau, 2014.Cho, 2014.Sutskever} offers numerous improvements and advantages in translation quality compared to traditional machine translation systems, such as statistical phrase-based systems \cite{2003.Koehn}. NMT also paved the way to translate multiple languages using a single model \cite{2017.Johnson}. Continued active research interest in the field of NMT has created many interesting architectures which produce models of high translation quality \cite{2017.Vaswani}. Recent research also shows how reduced precision and large batch training could speed-up the training while maintaining translation quality\cite{2018.Ott}.

There are several challenges when scaling out Deep Neural Network (DNN)-based models, such as efficiently exchanging gradients across multiple nodes, scaling up the batch size while maintaining generalized performance, and selecting appropriate hyper-parameters which efficiently train the model while preventing divergence and over-fitting. NMT approaches such as the transformer model \cite{2017.Vaswani}, which shares the weight matrix between the embedding layer and linear transformation before the \textit{softmax} layer, must ensure that the gradients from these two layers are updated appropriately without causing performance degradation or out-of-memory (OOM) errors.

In this paper, we begin by understanding the basics of a NMT model, and try to explore the reasons that restrict it's scalability. We then show how our current solution of forcibly densifying assumed-sparse tensors achieves high scaling efficiency -- both weak and strong -- when trained with up to 300 nodes on both the Zenith supercomputer at Dell EMC and the Stampede2 supercomputer at TACC. We also illustrate that even when trained with very large batch sizes (402k, 630k and 1 Million tokens), we are still able to achieve comparable or slightly better translation quality when compared to the official TensorFlow benchmark results.

The software changes which we discuss in this paper have been incorporated into Horovod 0.15.2 and later, providing other researchers the opportunity to apply this approach on any models that may benefit. 

\section{Background}
NMT models work much like source-to-source compilers, taking input from a source language (e.g., Fortran) and converting it to a target language (e.g., binary machine code). An NMT model first reads a sentence in a source language and passes it to an encoder, which builds an intermediate representation. This intermediate representation is then passed to the decoder, which processes the intermediate representation to produce the translated sentence in the target language.

\begin{figure}
    \centering
    \includegraphics[width=\columnwidth]{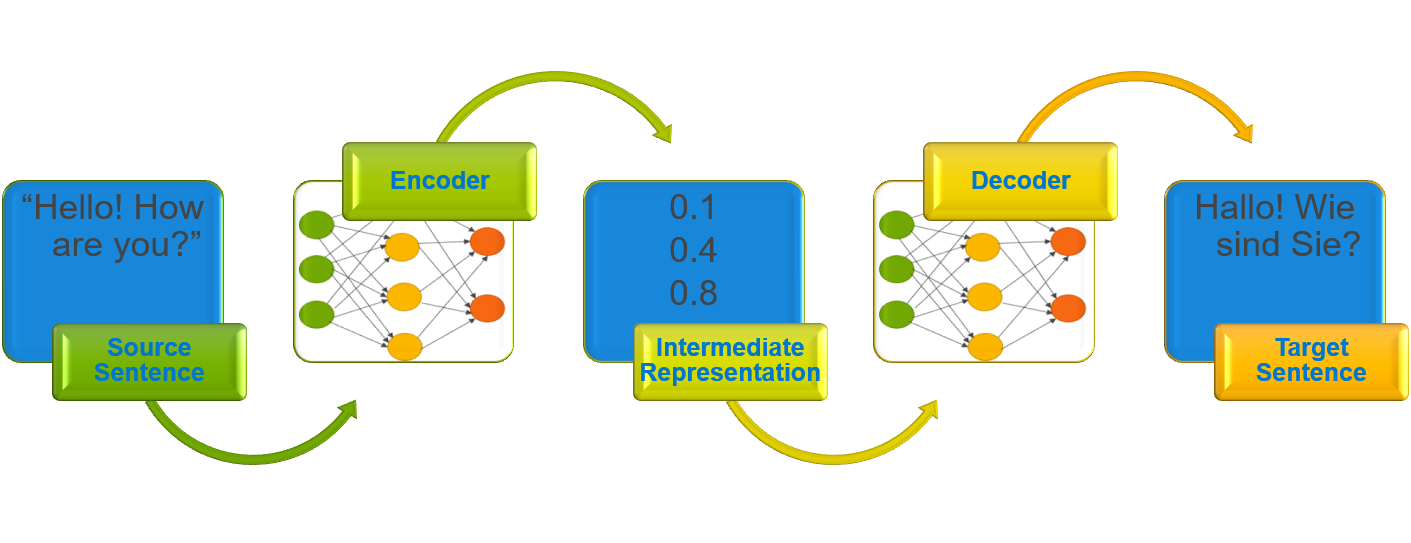}
    \caption{Encoder-decoder architecture}
    \label{fig:enc-dec}
\end{figure}

Fig.~\ref{fig:enc-dec} shows an encoder-decoder architecture. The English source sentence, ``Hello! How are you?''  is read and processed by the architecture to produce a translated German sentence ``Hallo! Wie sind Sie?''. Traditionally, Recurrent Neural Networks (RNN) were used in encoders and decoders~\cite{2014.Cho}, but other neural network architectures such as Convolutional Neural Networks (CNN)~\cite{2017.Gehring} and attention mechanism-based models~\cite{2015.Rush} are also used.

The transformer model~\cite{2017.Vaswani} is one of the interesting architectures in the field of NMT, which is built with variants of attention mechanism in the encoder-decoder part, eliminating the need for traditional RNNs in the architecture~\cite{2016.Collobert}. This model was able to achieve state of the art results in English-German and English-French translation tasks.

\begin{figure}
    \centering
    \includegraphics[width=.8\columnwidth]{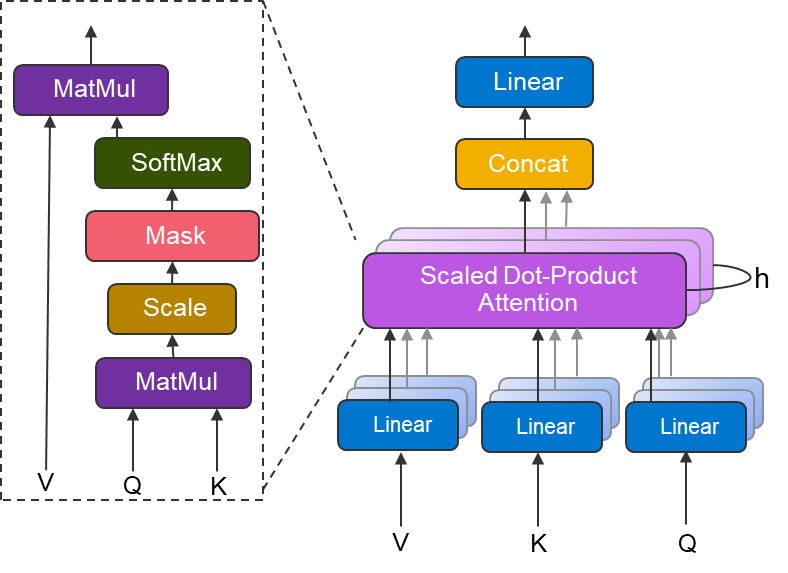}
    \caption{Multi-head attention block~\cite{2017.Vaswani}}
    \label{fig:attention}
\end{figure}

Fig.~\ref{fig:attention} illustrates the multi-head attention block used in the transformer model. At a high-level, the scaled dot-product attention can be imagined as finding the relevant information, values (V) based on Query (Q) and Keys (K) and multi-head attention could be thought as several attention layers in parallel to get distinct aspects of the input.

\section{Issues with Scaling the Transformer Model}

Encoder-decoder models for NMT make use of an attention mechanism to help the decoders obtain the vital information from the source tokens while discarding irrelevant information. The main structure of the transformer model is the multi-head attention, which provides a way to get different linear transformations of all the inputs. These components allow an NMT model to learn more robustly.  But a particular design consideration that needs to be looked at for improving the scaling capabilities is the weight matrix that is shared between the embedding layer and the projection matrix. This type of similar design is also seen in other NMT models such as ~\cite{2017.Gehring}. Hence, understanding the cause and effect of these specific design considerations is vital for the NMT research community.

This particular design would cause performance degradation or OOM errors if the gradients from these layers are not accumulated correctly. Specifically, gradients from the embedding layer are sparse whereas the gradients from the projection matrix are dense. In TensorFlow both gradients are updated together as a sparse \verb$IndexedSlices$ objects. This has a dramatic effect on TensorFlow's determination of a gradient accumulation strategy, and subsequently on the total size of the accumulated gradient tensor. 

\begin{algorithm}
\caption{Tensor Accumulation Strategy in TensorFlow~\cite{TFAccumulate}}
\label{alg:accumulate}
\begin{algorithmic}[1]
  \If {$\lvert GRAD_{in}\rvert<2$}
    \State $GRAD_{out}\gets GRAD_{in}$ \Comment Pass-through  
  \ElsIf {$type(g) = Tensor~\forall g\in GRAD_{in}$}
    \State $GRAD_{out}\gets \sum GRAD_{in}$ \Comment Output is a dense Tensor (reduce)
  \Else
    \State $GRAD_{out}\gets \overset{\frown}{GRAD_{in}}$ \Comment Output is a sparse IndexedSlice (gather)
  \EndIf
\end{algorithmic}
\end{algorithm}

Algorithm~\ref{alg:accumulate} describes the algorithm used in TensorFlow to accumulate gradients, based on the assumed type and shape of the gradients being accumulated (see~\cite{TFAccumulate}). At present, TensorFlow will either: (1) do nothing if there are less than 2 output gradients, (2) accumulate gradients by reduction if all gradients are expressed as dense tensors with defined shapes, or (3) convert everything to indexed slices and accumulate by concatenation (performing a gather operation).

In this particular use case, the embedding lookup is performed using \\\verb$tf.gather$, which returns an \verb$IndexedSlice$ object. This forces TensorFlow (based on the accumulation algorithm - Algorithm~\ref{alg:accumulate}) to convert the remaining dense tensors to indexed slices, even though all the gradients being accumulated are dense.

\begin{figure}
    \centering
    \subfloat[Before: tf.gather/MPI\_Gather\label{fig:oom}]{
      \includegraphics[width=.7\columnwidth]{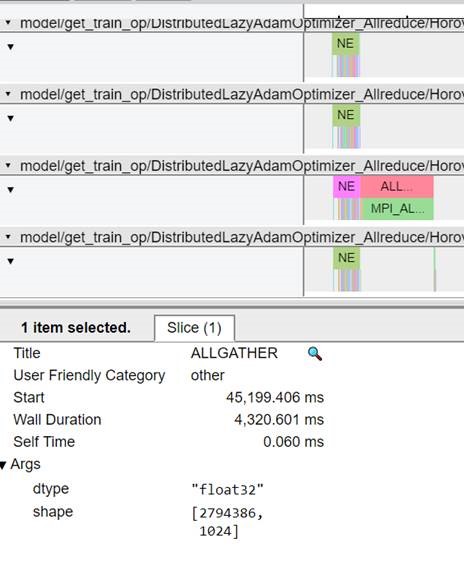}
    }
    \\
    \subfloat[After: tf.reduce/MPI\_Reduce\label{fig:after}]{
      \includegraphics[width=.7\columnwidth]{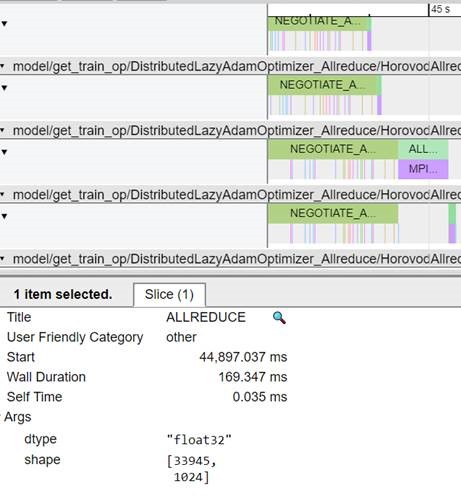}
    }
    \caption{Horovod timelines for 64 MPI process tests before and after modification}
    \label{fig:hvd-timeline}
\end{figure}

The result of this decision to convert and assume that the gradient tensors are sparse is to accumulate by gathering, rather than reduction. This applies not only to single-node tensor accumulation, but to multi-node accumulation through Horovod due to the use of the main TensorFlow graph in determining which collective operations Horovod will perform using MPI. The result is extremely large message buffers (exceeding 11GB - see Fig.~\ref{fig:oom}), which cause segmentation faults or out-of-memory (OOM) errors.

\begin{figure}
    \centering
    \includegraphics[width=.7\columnwidth]{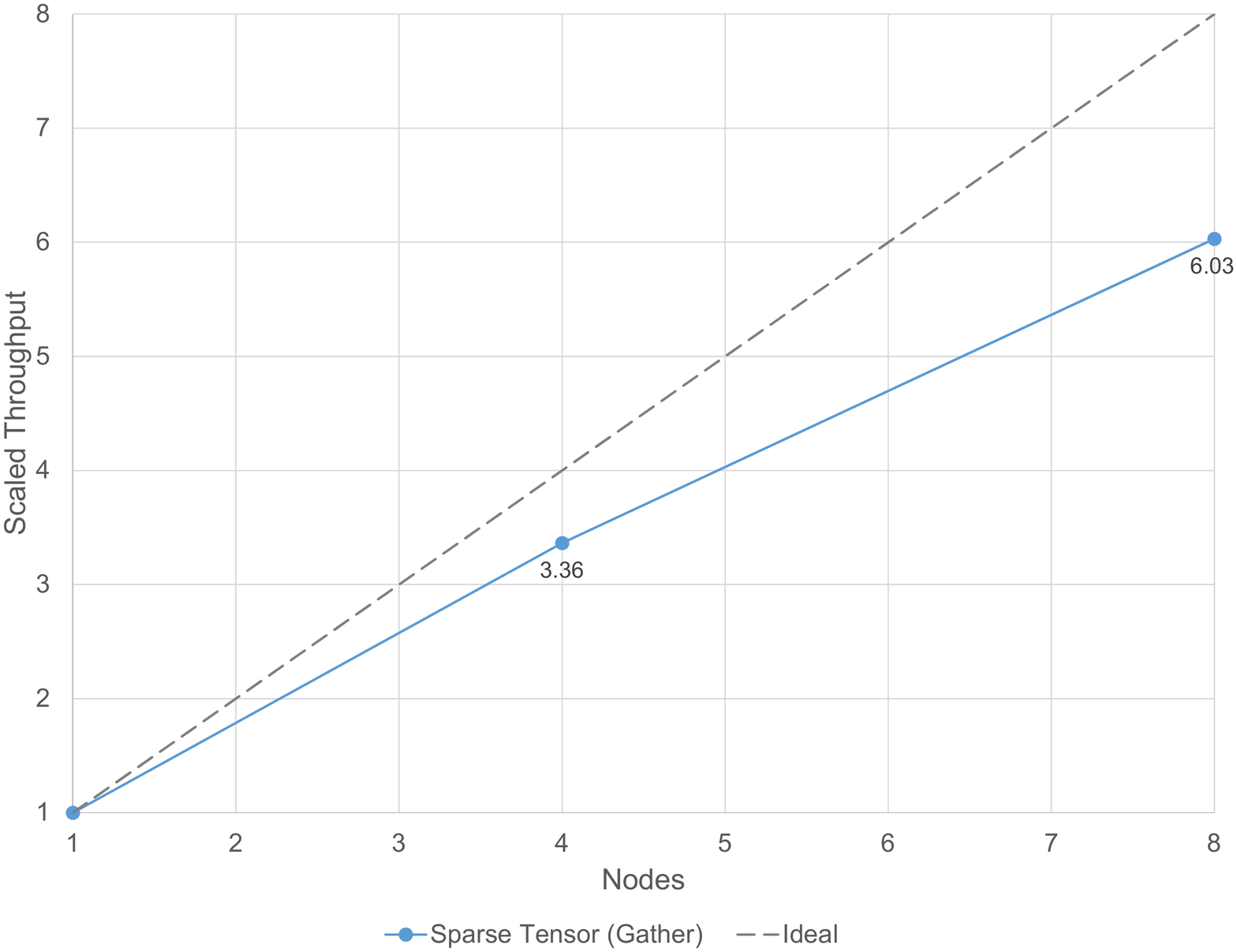}
    \caption{Scaled speedup with sparse tensor accumulation strategy (gather)}
    \label{fig:scaling_zenith_before}
\end{figure}

Because of the message buffer sizes, we were unable to scale beyond 32 MPI processes, and saw quickly diminishing scaling efficiency, or fraction of ideal scaled speedup. Fig.~\ref{fig:scaling_zenith_before} shows the scaled speedup of the training process up to the maximum achievable 32 MPI processes (8 nodes with 4 processes per node). Scaling efficiency -- which is visually expressed as distance from the ideal line -- declines rapidly, going from 84\% with 4 nodes to 75\% for 8 nodes. Eventually scaled speedup would (if the training could be parallelized further) reach an asymptotic limit where additional resources do not further accelerate the algorithm.

\section{Densifying Assumed-sparse Tensors}

In order to correct for the issue of assumed-sparse tensors in TensorFlow, we have implemented a forced-conversion of all gradient tensors to dense representation inside of Horovod's \verb$DistributedOptimizer$ method. This will then force TensorFlow to accumulate those tensors via reduction, rather than aggregation (see Listing~\ref{lst:hvd}).

\begin{lstlisting}[float,language=Python, basicstyle=\ttfamily, caption={Horovod code for converting IndexedSlices to Tensors~\cite{HvdFix}\label{lst:hvd}}]
for grad, var in gradients:
   if grad is not None:
      if self._sparse_as_dense and 
         isinstance(grad, tf.IndexedSlices):
         grad = tf.convert_to_tensor(grad)
\end{lstlisting}

\begin{figure}
    \centering
    \subfloat[Accumulate size (MB)]{
      \includegraphics[width=.48\columnwidth]{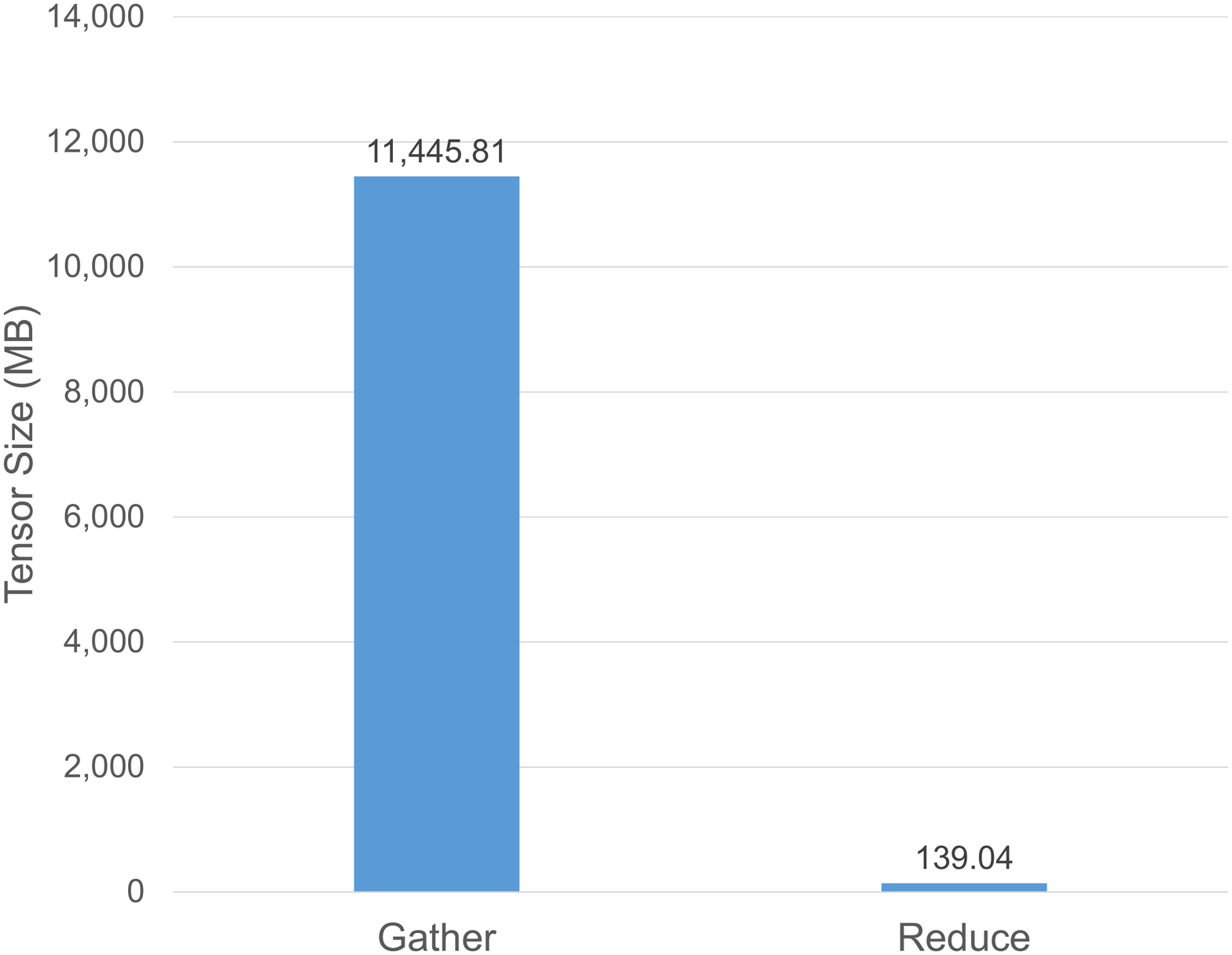}
    }
    \hfill
    \subfloat[Accumulate time (ms)]{
      \includegraphics[width=.48\columnwidth]{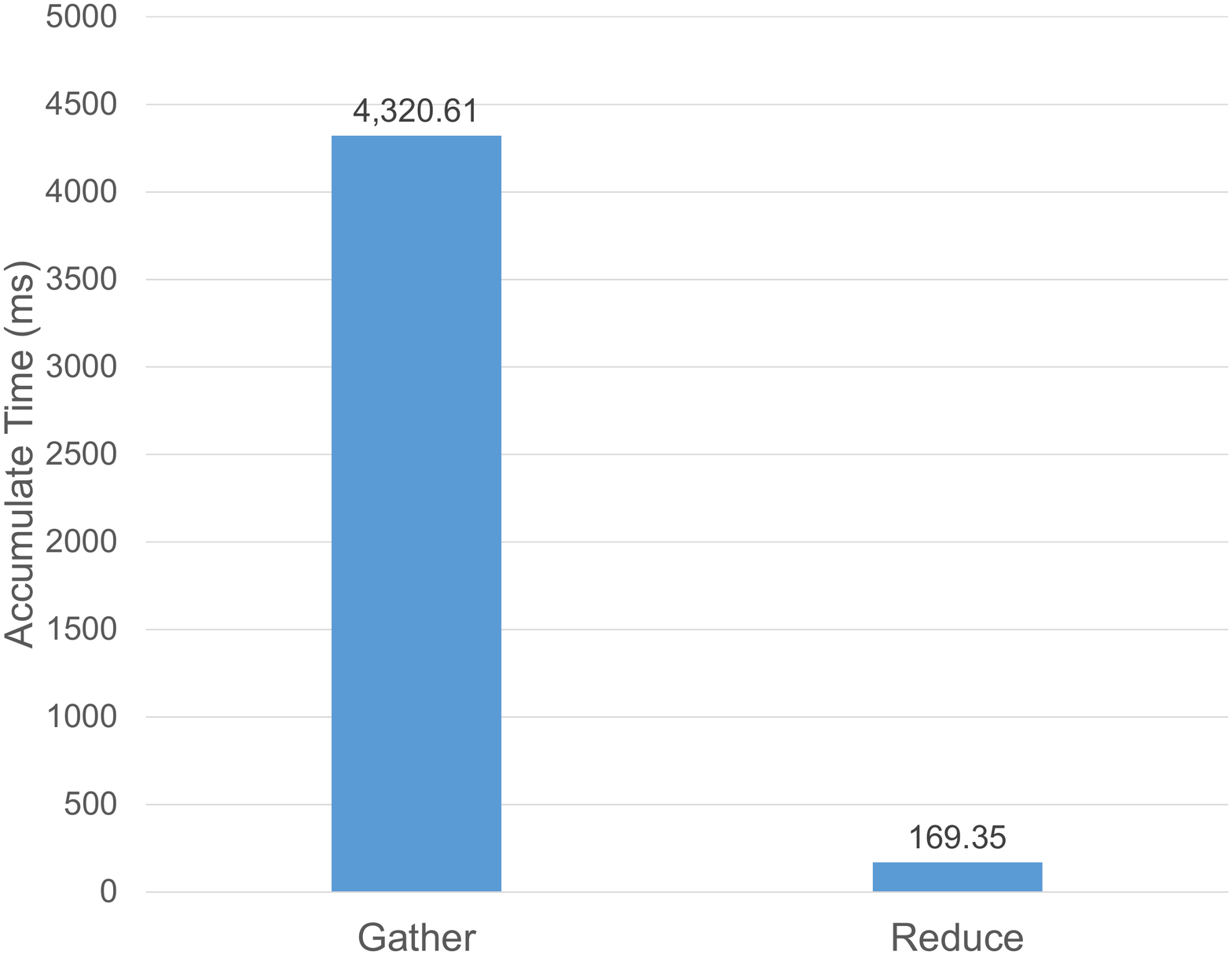}
    }
    \caption{Space/time for tensor accumulate (sparse gather vs. dense reduce)}
    \label{fig:accumulate}
\end{figure}

The result is an 82x reduction in the amount of memory required (from 11.4GB to 139MB - see Fig.~\ref{fig:oom} and Fig.~\ref{fig:after}, respectively) when using 64 nodes (1 MPI process per node, batch size 5000 tokens). Additionally, the time needed to perform the accumulate operation drops from 4320 ms to 169 ms, which is a 25x reduction (see Fig.~\ref{fig:accumulate} for a comparison of accumulate size and time).

These small changes reduce the memory footprint per process to a degree that we can both scale up the batch size per MPI process and increase the number of MPI processes per run. They also reduce the tensor exchange time significantly enough to maintain near-linear scaling when running in a multi-node environment.

This algorithmic change can be made in Horovod 0.15.2 or later by setting the \verb$sparse_as_dense$ option when initializing \verb$DistributedOptimizer$:
\begin{verbatim}
    opt = hvd.DistributedOptimizer(opt, sparse_as_dense=True)
\end{verbatim}

\section{Experimental Results}
The models were trained using the WMT-17 English-German parallel corpus with 4.5M sentence pairs. The newstest2014 dataset was used as unseen test data to capture the translation quality. All the pre-processing and BLEU~\cite{papineni2002bleu} calculations were in accordance with TensorFlow's official benchmarks in order to compare performance and translation quality. We also used hyper parameter settings based on best practices in~\cite{2018.Popel, 2018.Ott}. Model training experiments were run on the Zenith cluster in the Dell EMC HPC \& AI Innovation Lab, as well as the Stampede2 cluster at the Texas Advanced Computing Center (TACC) in Austin, Texas.

Each Zenith node contains dual Intel\textsuperscript\textregistered Xeon\textsuperscript\textregistered Scalable Gold 6148/F processors, 192GB of memory, and an M.2 boot drive to house the operating system that does not provide user-accessible local storage. Nodes are interconnected by a 100Gbps Intel\textsuperscript\textregistered Omni-path fabric, and shared storage is provided by a combination of NFS (for \texttt{HOME} directories) and Lustre~\cite{schwan2003lustre} filesystems.

For our Zenith tests, we used Python 2.7, with Intel's MKL-optimized version of TensorFlow (1.12). The version of Horovod used for these experiments was a private branch for testing purposes, but all of these optimizations have now been made a part of Horovod 0.15.2. Table~\ref{tab:zenith_sw} gives a complete breakdown of the software environment used for the Zenith experiments, while Listing~\ref{lst:zenith_runtime} provides the runtime settings for the experiments.

\begin{table}[]
    \centering
    \begin{tabular}{|r||l|}
         \hline
         \bf Package & \bf Version\\
         \hline
         \hline
         Python &  2.7.13\\
         \hline
         TensorFlow & Anaconda TensorFlow 1.12.0 with Intel\textsuperscript\textregistered MKL\\
         \hline
         Horovod & 0.15.2\\
         \hline
         MPI & MVAPICH2 2.1\\
         \hline
    \end{tabular}
    \caption{Software Environment for Zenith Experiments}
    \label{tab:zenith_sw}
\end{table}

\begin{lstlisting}[float,language=bash,caption={Runtime settings for Zenith Experiments\label{lst:zenith_runtime}},basicstyle=\ttfamily]
OMP_NUM_THREADS=10
KMP_BLOCKTIME=0
KMP_AFFINITY=granularity=fine,verbose,compact,1,0
HOROVOD_FUSION_THRESHOLD=134217728
\end{lstlisting}

We also ran scaling tests on the Stampede2 cluster at the Texas Advanced Computing Center (TACC) at The University of Texas at Austin~\cite{Stanzione:2017:3093338.3093385}. Stampede2 has two partitions, each with a different set of processors. Our tests were performed on the SKX partition, which consists of 1,736 nodes, each with dual Intel\textsuperscript\textregistered Xeon\textsuperscript\textregistered Scalable Platinum 8160 processors, 192GB of memory, and 200GB internal SSD drive for the operating system and local \texttt{/tmp}. The second KNL partition consists of 4,200 nodes, each with a single Intel\textsuperscript\textregistered Xeon Phi\textsuperscript{TM} 7250 processor with 16GB of on-package MCDRAM, 94GB of main memory, and a 200GB SSD for the operating system and local \texttt{/tmp}. All nodes are interconnected with 100Gbps Intel\textsuperscript\textregistered Omni-path fabric and connected to Lustre-based shared filesystems.

For our Stampede2 tests, we used Python 2.7, with Intel's MKL-optimized version of TensorFlow (1.12). The version of Horovod used for these experiments was a private branch for testing purposes, but all of these optimizations have now been made a part of Horovod 0.15.2. Table~\ref{tab:stampede2_sw} gives a complete breakdown of the software environment used for the Zenith experiments.

\begin{table}[]
    \centering
    \begin{tabular}{|r||l|}
         \hline
         \bf Package & \bf Version\\
         \hline
         \hline
         Python &  2.7.13\\
         \hline
         TensorFlow & Anaconda TensorFlow 1.12.0 with Intel\textsuperscript\textregistered MKL\\
         \hline
         Horovod & 0.15.2\\
         \hline
         MPI & MVAPICH2 2.3\\
         \hline
    \end{tabular}
    \caption{Software Environment for Stampede2 Experiments}
    \label{tab:stampede2_sw}
\end{table}

\subsection{Weak Scaling Performance}
\begin{figure}
    \centering
    \includegraphics[width=.7\columnwidth]{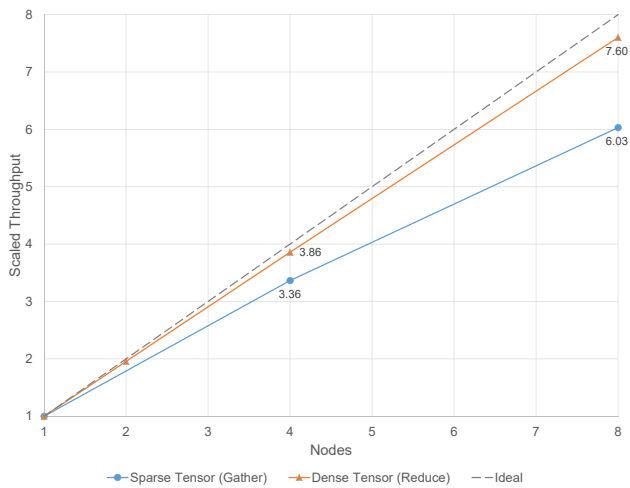}
    \caption{Comparison of weak scaling on Zenith up to 8 nodes (4PPN) between sparse and dense tensor accumulation strategies}
    \label{fig:zenith_compare}
\end{figure}

The difference in reducing the output gradient size can be seen when comparing the scaling efficiency -- the ratio between observed scaled speedup and ideal -- between the default sparse tensor accumulation strategy (gather) and the dense tensor accumulation strategy (reduce). Dense tensor accumulations show significantly better scaling efficiency out to 32 MPI processes (95\%) than the default sparse tensor accumulation (75\%) (see Fig.~\ref{fig:zenith_compare}).

\begin{figure}
    \centering
    \subfloat[Throughput (tokens/second)]{
      \includegraphics[width=.48\columnwidth]{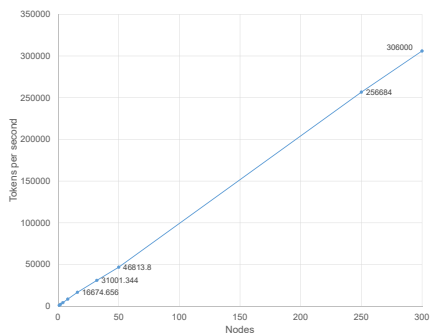}
    }
    \hfill
    \subfloat[Scaled throughput\label{fig:throughput_zenith}]{
      \includegraphics[width=.48\linewidth]{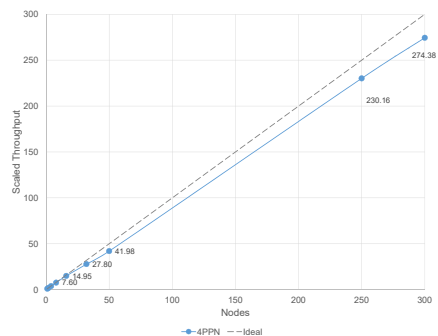} 
    }
    \caption{Weak scaling on Zenith cluster from 1 to 300 nodes (4 PPN) using dense tensor accumulation strategy (reduce)}
    \label{fig:weak_zenith}
\end{figure}

\begin{figure}
    \centering
    \includegraphics[width=.7\columnwidth]{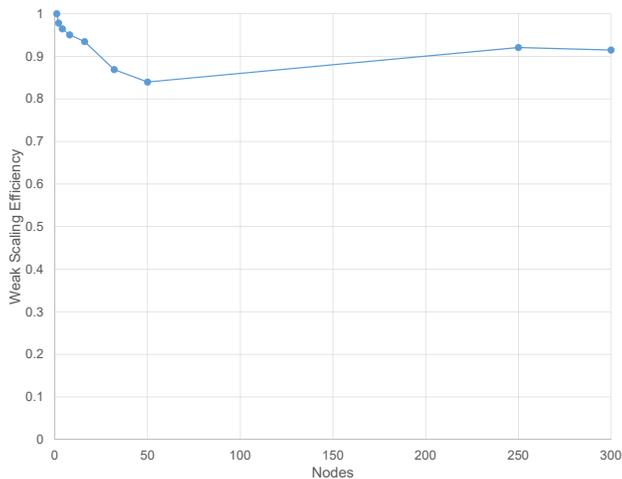}
    \caption{Weak scaling efficiency on Zenith up to 300 nodes (1200 processes)}
    \label{fig:weak_efficiency_zenith}
\end{figure}

The reduced output gradient size and improved scaling efficiency mean that we can scale to larger process counts than was previously possible. Additional weak scaling experiments on Zenith using 4 processes per node (PPN) on up to 300 compute nodes (1200 MPI processes) show near-linear scaling, with efficiency dropping from 95\% for 8 nodes to 91.5\% for 300 (see Fig.~\ref{fig:weak_zenith} and Fig.~\ref{fig:weak_efficiency_zenith}). For these particular experiments on Zenith, batch size per process was held constant at 5000 tokens, or 20000 tokens per node. This means in the largest case (1200 MPI processes) we are training with a global batch size of 6M tokens. 

The ability to maintain very high weak scaling efficiency above 90\% suggests that continued scale-out is worthwhile. We will seek to perform additional experiments on systems larger than Zenith.

\subsection{Strong Scaling}
Besides good weak scaling efficiency, the reduced output gradient size also gives us the possibility to perform strong scaling experiments. For this purpose, we have selected a global batch size of 819,200 that allows us to produce a near-state-of-the-art model in terms of translation quality (as measured by BLEU score~\cite{papineni2002bleu}), and as discussed in the following section. Obtaining good strong scaling efficiency is significantly more challenging compared to the weak scaling case, as the effective batch size per worker decreases when increasing the node count.

We have performed strong scaling experiments on both on the Zenith cluster and on the Stampede2 supercomputer from TACC. We have used up to 200 nodes on Zenith, and up to 512 nodes on Stampede2, both systems showing significant reductions in terms of time to solution.

\begin{figure}
    \centering
    \includegraphics[width=.7\columnwidth]{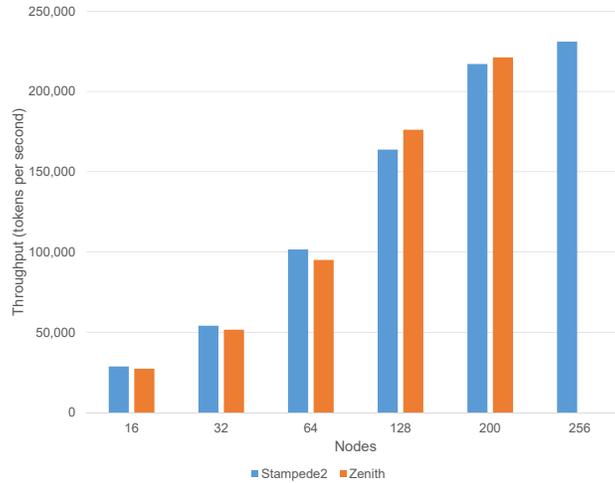}
    \caption{Strong scaling throughput on up to 200 nodes of Zenith (Dell EMC) and 256 nodes of Stampede2 (TACC) with global batch size of 819,200 tokens}
    \label{fig:throughput_cmp}
\end{figure}

\begin{figure}
    \centering
    \includegraphics[width=.7\columnwidth]{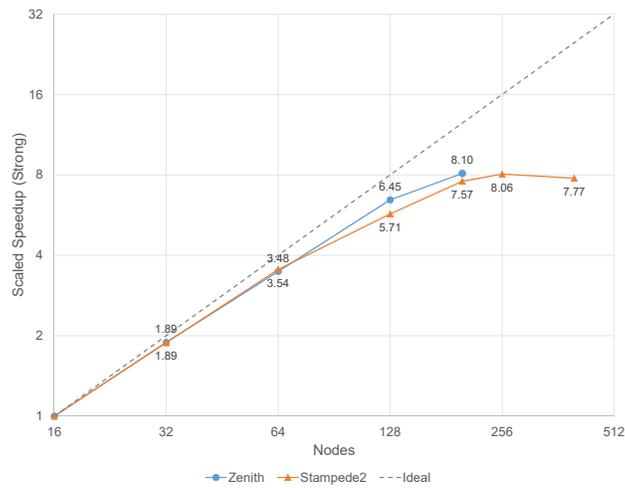}
    \caption{Scaled speedup (strong scaling) up to 200 nodes on Zenith (Dell EMC) and 256 nodes on Stampede2 (TACC) with a global batch size of 819,200 tokens}
    \label{fig:strong_efficiency_zenith}
\end{figure}

Fig.~\ref{fig:strong_efficiency_zenith} and Fig.~\ref{fig:throughput_cmp} illustrate the strong scaling behavior that can be expected on the Zenith system. When going from 16 nodes up to 200 nodes, we can improve the throughput by a factor exceeding 8 (out of a maximum of around 12). In all these strong scaling cases, we only use 2 processes per node, each being scheduled to run on one socket and exploiting the NUMA affinity. This setting is more appropriate in this scenario, as the batch size that can be used per worker is double compared to the case when using 4 processes per node.

\begin{figure}
    \centering
    \includegraphics[width=.7\columnwidth]{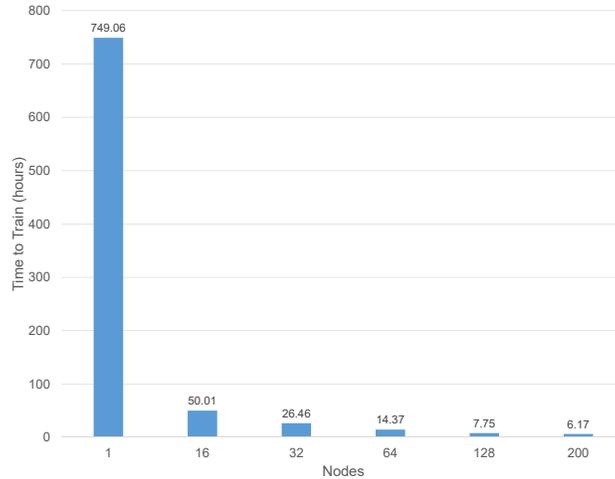}
    \caption{Time to solution (strong scaling) on up to 200 Zenith nodes}
    \label{fig:strong_zenith_tts}
\end{figure}

The impact of having good strong scaling efficiency is that training times can be dramatically reduced. This can be best visualized in Fig.~\ref{fig:strong_zenith_tts}, where the time to solution drops from around one month when using a single node, down to slightly over 6 hours when using 200 nodes (121 times faster), therefore significantly increasing the productivity for NMT researchers when using CPU-based HPC infrastructures. The results observed were based on the models achieving a baseline BLEU score (case-sensitive) of 27.5.

For the single node case, we have used the largest batch size that could fit in a node's memory, 25,600 tokens per worker. For all other cases we use a global batch size of 819,200, leading to per-worker batch sizes of 25,600 in the 16-node case, down to only 2,048 in the 200-node case. The number of training iterations is similar for all experiments in the 16-200 node range, and is increased by a factor of 16 for the single-node case (to compensate for the larger batch).

On Stampede2, the behavior is similar to zenith up to 200 nodes. Since Stampede2 is a larger system, we performed larger strong scaling experiments. However, we noticed that using a 819,200 batch size would limit the scaling efficiency when using over 256 nodes. The 200 to 256 node range show improvements in time-to-solution, but when using 400 nodes we have reached the limits of strong scaling, and begin to observe performance degradation. This is due to the fact that a small (1,024) per-worker batch size is used in the 400 nodes experiment. To test that this is the case, we performed a larger experiment using a per-worker batch size of 1,536, and a total of 1,024 workers divided across 512 nodes. This leads to a global batch size of 1,572,864, and requires further attention to in order to reach the translation accuracy performance targets.  However, from a throughput perspective, this run is 56\% faster compared to a similar 256-node run. This shows that there will be performance improvements as we increase the per-worker batch size to a reasonably large size ($> 1536$).

\subsection{Model Accuracy}
\begin{figure}
    \centering
    \includegraphics[width=.7\columnwidth]{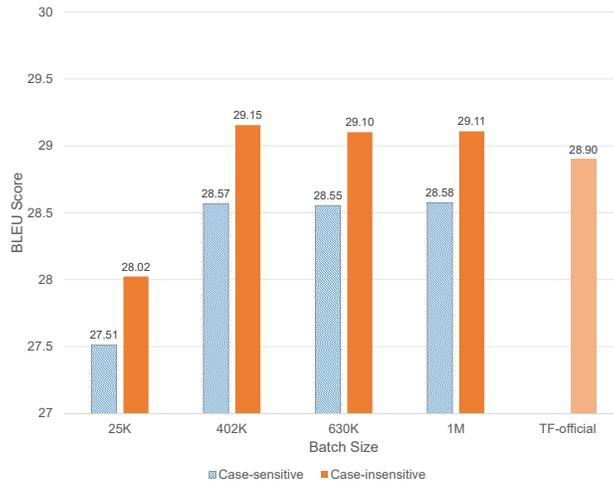}
    \caption{Translation quality (BLEU) when trained with varying batch size on Zenith}
    \label{fig:zenith_accuracy}
\end{figure}

Scaling out transformer model training using MPI and Horovod improves throughput performance, while producing models of similar translation quality (see Fig.~\ref{fig:zenith_accuracy}). Models of comparable quality can be trained in a reduced amount of time by scaling computation over many more nodes, and with larger global batch sizes (GBZ). Our experiments on Zenith demonstrate ability to train models of comparable or higher translation quality (as measured by BLEU score~\cite{papineni2002bleu}) than the reported best for TensorFlow's official model~\cite{TFTransformer}, even when training with batches of a million or more tokens.

\section{Discussion}
Our experiments have demonstrated that converting assumed-sparse tensors to dense tensors improves memory utilization as well as time to accumulate, thanks to a switch from gathering to reduction (see Fig.~\ref{fig:accumulate}). Unlike similar solutions implemented directly within optimized NMT models, such as NVIDIA's OpenSeq2Seq package~\cite{2018.Kuchaiev}, our approach \emph{does not limit usability strictly to one specific package repository or model implementation}. Instead, our approach provides greater generalized use and potential applicability to other models. 

\paragraph{Applicability to other Models.}
We believe the solution that is now implemented in Horovod will prove useful to most neural network model classes, including various language translation models, image segmentation models, voice/text translation models across multiple voice datasets, time-series models, etc. Future work will quantify the impact of the current solution to these use cases. We also foresee this as a potential workaround for issues in custom architectures, such as multi-branch neural networks~\cite{xie2017aggregated,yamashita2005application,yamashita2002multi,hu2018squeeze}. These architectures are typically recollecting gradient data from multiple ``separated'' neural network branches, which would be likely to encounter similar sparse tensor encoding issues.

\paragraph{Specificity to TensorFlow.}
While we have identified a specific edge case within the TensorFlow code base, we do not believe that this particular edge case is common to other deep learning frameworks, such as Caff\'e2~\cite{2014.Jia} and PyTorch~\cite{2017.Paszke}. However, TensorFlow's current and continuing popularity and the abundance of pre-built models in TensorFlow mean that any performance benefits we can communicate back to that community are important.

\paragraph{Incorporating Changes into TensorFlow.}
Long-term, we believe that the ideal solution is to add additional logic into TensorFlow's gradient accumulation algorithm to convert and reduce tensors when \textit{\textbf{any}} of the tensors is dense (see Algorithm~\ref{alg:new_accumulate}), rather than only when \textit{\textbf{all}} of the tensors are dense (as is the case in Algorithm~\ref{alg:accumulate}).

\begin{algorithm}
\caption{Proposed Tensor Accumulation Strategy for TensorFlow}
\label{alg:new_accumulate}
\begin{algorithmic}[1]
  \If {$\lvert GRAD_{in}\rvert<2$}
    \State $GRAD_{out}\gets GRAD_{in}$ \Comment Pass-through  
  \ElsIf {$type(g) = Tensor~\forall g\in GRAD_{in}$}
    \State $GRAD_{out}\gets \sum GRAD_{in}$ \Comment Output is a dense Tensor (reduce)
  \ElsIf {$\exists g\in GRAD_{in}~type(g)=Tensor$}
    \State $GRAD_{conv} \gets \{conv\_to\_tensor(g),~\forall g\in GRAD_{in}\}$ \Comment Convert all to Tensor
    \State $GRAD_{out}\gets\sum GRAD_{conv}$ \Comment Output is a dense  Tensor (reduce)
  \Else
    \State $GRAD_{out}\gets \overset{\frown}{GRAD_{in}}$ \Comment Output is a sparse IndexedSlice (gather)
  \EndIf
\end{algorithmic}
\end{algorithm}

In the case of Algorithm~\ref{alg:new_accumulate}, we propose the addition of an extra conditional block (lines 5--7), which would handle the case that there exists at least 1 tensor which is dense, in which case all of the tensors to be accumulated would be converted to dense and accumulated by reduction. More research has to be done in order to ensure that incorporating this conditional block into the TensorFlow accumulation strategy would not adversely effect other well-behaved tensor accumulations, and we will be testing this inclusion and proposing back to TensorFlow in the future.

\section{Future Work \& Conclusion}
Scaling Neural Machine Translation (NMT) models to multiple nodes can be difficult due to the large corpuses needed for reasonable translation, and the all-to-all mapping nature of the intermediate representation encodings. If tensor accumulation is not performed in a memory and compute-optimized fashion, excessively large tensors can cause buffer overruns which prevent scaling beyond a few MPI processes. These models can take weeks or months to train at low node counts, making it all the more critical that they can be efficiently scaled to hundreds or thousands of MPI processes.

We have identified an edge case in TensorFlow's tensor accumulation strategy which leads to sub-optimal memory and compute utilization, which prevents scaling of multi-head attention-based transformer models beyond a relatively small number of processes without very large memory buffers. We have proposed and implemented a fix via the Horovod MPI-based framework for distributed memory scaling of TensorFlow models by forcibly converting -- through the use of an option to \verb$DistributedOptimizer$ -- all tensors to be accumulated to dense and subsequently reducing tensors rather than aggregating them. The result is a more than 82x reduction in memory needed and 25x reduction in time to complete the accumulation step at 64 MPI processes, and the enabled ability to scale the translation model to a thousand MPI processes or more with batches of millions of word part tokens.

These modifications have been incorporated into Horovod, and are available as of version 0.15.2~\cite{HvdTag}, so that other teams can scale neural machine translation tasks or any other tasks which use similar topologies. We have proposed a potential fix within TensorFlow as  a more long-term solution to this issue, and we will be pursuing this going forward once we have determined that there are no additional side-effects from the addition of the new tensor accumulation strategy.

Going forward, we intend to investigate whether other neural network architectures besides multi-head attention can benefit from being able to expressly densify sparse tensor encodings, as well as whether custom architectures could potentially benefit from this solution.

\section*{Acknowledgement}
The authors acknowledge the Texas Advanced Computing Center (TACC) at The University of Texas at Austin for providing HPC resources that have contributed to the research results reported within this paper.  \url{http://www.tacc.utexas.edu} 

\bibliographystyle{plain}
\bibliography{00_dondsmith2,supplemental_refs}

\begin{thebibliography}{10}

\bibitem{2014.Bahdanau}
Dzmitry Bahdanau, Kyunghyun Cho, and Yoshua Bengio.
\newblock {Neural Machine Translation by Jointly Learning to Align and
  Translate}.
\newblock {\em Computing Research Repository ({CoRR})}, abs/11409.0473v7,
  September 2014.

\bibitem{2014.Cho}
Kyunghyun Cho, Bart van Merrienboer, Caglar Gulcehre, Fethi Bougares, Holger
  Schwenk, and Yoshua Bengio.
\newblock {Learning Phrase Representations using RNN Encoder-Decoder for
  Statistical Machine Translation}.
\newblock {\em Computing Research Repository ({CoRR})}, abs/1406.1078v3,
  September 2014.

\bibitem{2016.Collobert}
Ronan Collobert, Christian Puhrsch, and Gabriel Synnaeve.
\newblock {Wav2Letter: an End-to-End ConvNet-based Speech Recognition System}.
\newblock {\em Computing Research Repository ({CoRR})}, abs/1609.03193v2,
  September 2016.

\bibitem{2017.Gehring}
Jonas Gehring et~al.
\newblock {Convolutional Sequence to Sequence Learning}.
\newblock {\em Computing Research Repository ({CoRR})}, abs/1705.03122v3, July
  2017.

\bibitem{HvdFix}
Horovod.
\newblock {compute\_gradients() in horovod/tensorflow/\_\_init\_\_.py}.
\newblock
  \url{https://github.com/uber/horovod/blob/085cb1b5f3b30734a34d047841b098c15a6e1bae/horovod/tensorflow/__init__.py#L195}.

\bibitem{HvdTag}
Horovod.
\newblock {Release 0.15.2}.
\newblock \url{https://github.com/uber/horovod/releases/tag/v0.15.2}.

\bibitem{hu2018squeeze}
Jie Hu, Li~Shen, and Gang Sun.
\newblock Squeeze-and-excitation networks.
\newblock In {\em Proceedings of the IEEE conference on computer vision and
  pattern recognition}, pages 7132--7141, 2018.

\bibitem{2014.Jia}
Yangqing Jia, Evan Shelhamer, Jeff Donahue, Sergey Karayev, Jonathan Long, Ross
  Girshick, Sergio Guadarrama, and Trevor Darrell.
\newblock {Caffe: Convolutional Architecture for Fast Feature Embedding}.
\newblock In {\em {Proceedings of the 22Nd ACM International Conference on
  Multimedia}}, pages 675--678, November 2014.

\bibitem{2017.Johnson}
Melvin Johnson, Mike Schuster, Quoc~V Le, Maxim Krikun, Yonghui Wu, Zhifeng
  Chen, Nikhil Thorat, Fernanda Viegas, Martin Wattenberg, Greg Corrado,
  Macduff Hughes, and Jeffrey Dean.
\newblock {Google's Multilingual Neural Machine Translation System: Enabling
  Zero-Shot Translation}.
\newblock {\em Computing Research Repository ({CoRR})}, abs/1611.04558v2,
  August 2017.

\bibitem{2003.Koehn}
Philipp Koehn, Franz~Josef Och, and Daniel Marcu.
\newblock {Statistical Phrase-Based Translation}.
\newblock In {\em {Proceedings of 2003 Human Language Technology Conference
  (HLT-NAACL)}}, pages 48--54, June 2003.

\bibitem{2018.Kuchaiev}
Oleksii Kuchaiev, Boris Ginsburg, Igor Gitman, Vitaly Lavrukhin, Carl Case, and
  Paulius Micikevicius.
\newblock {Mixed-Precision Training for NLP and Speech Recognition with
  OpenSeq2Seq}.
\newblock {\em Computing Research Repository ({CoRR})}, abs/1805.10387v2,
  November 2018.

\bibitem{2018.Ott}
Myle Ott, Sergey Edunov, David Grangier, and Michael Auli.
\newblock {Scaling Neural Machine Translation}.
\newblock {\em Computing Research Repository ({CoRR})}, abs/1806.00187v3,
  September 2018.

\bibitem{papineni2002bleu}
Kishore Papineni, Salim Roukos, Todd Ward, and Wei-Jing Zhu.
\newblock Bleu: a method for automatic evaluation of machine translation.
\newblock In {\em Proceedings of the 40th annual meeting on association for
  computational linguistics}, pages 311--318. Association for Computational
  Linguistics, 2002.

\bibitem{2017.Paszke}
Adam Paszke, Sam Gross, Soumith Chintala, Gregory Chanan, Edward Yang, Zachary
  DeVito, Zeming Lin, Alban Desmaison, Luca Antiga, and Adam Lerer.
\newblock {Automatic Differentiation in PyTorch}, December 2017.

\bibitem{2018.Popel}
Martin Popel and Ondrej Bojar.
\newblock {Training Tips for the Transformer Model}.
\newblock {\em Computing Research Repository ({CoRR})}, abs/1804.00247v2, May
  2018.

\bibitem{2015.Rush}
Alexander~M Rush, Sumit Chopra, and Jason Weston.
\newblock {A Neural Attention Model for Abstractive Sentence Summarization}.
\newblock {\em Computing Research Repository ({CoRR})}, abs/1509.00685v2,
  September 2015.

\bibitem{schwan2003lustre}
Philip Schwan et~al.
\newblock Lustre: Building a file system for 1000-node clusters.
\newblock In {\em Proceedings of the 2003 Linux symposium}, volume 2003, pages
  380--386, 2003.

\bibitem{Stanzione:2017:3093338.3093385}
Dan Stanzione, Bill Barth, Niall Gaffney, Kelly Gaither, Chris Hempel, Tommy
  Minyard, S.~Mehringer, Eric Wernert, H.~Tufo, D.~Panda, and P.~Teller.
\newblock Stampede 2: The evolution of an xsede supercomputer.
\newblock In {\em Proceedings of the Practice and Experience in Advanced
  Research Computing 2017 on Sustainability, Success and Impact}, PEARC17,
  pages 15:1--15:8, New York, NY, USA, 2017. ACM.

\bibitem{2014.Sutskever}
Ilya Sutskever, Oriol Vinyals, and Quoc~V Le.
\newblock {Sequence to Sequence Learning with Neural Networks}.
\newblock In {\em Advances in Neural Information Processing Systems 27}, NIPS
  Proceedings, pages 3104–--3112, December 2014.

\bibitem{TFAccumulate}
TensorFlow.
\newblock {\_AggregatedGrads() in tensorflow/python/ops/gradients\_impl.py}.
\newblock
  \url{https://github.com/tensorflow/tensorflow/blob/c95ca05536144451ef78ca6e2c15f0f65ebaaf95/tensorflow/python/ops/gradients_impl.py#L1183}.

\bibitem{TFTransformer}
TensorFlow.
\newblock {Official Transformer Model}.
\newblock
  \url{https://github.com/tensorflow/models/blob/cdcd3ec276bdccd77a9a35c38f5aaec39c15cc0b/official/transformer/README.md}.

\bibitem{2017.Vaswani}
Ashish Vaswani et~al.
\newblock {Attention Is All You Need}.
\newblock {\em Computing Research Repository ({CoRR})}, abs/1706.03762v5,
  December 2017.

\bibitem{xie2017aggregated}
Saining Xie, Ross Girshick, Piotr Doll{\'a}r, Zhuowen Tu, and Kaiming He.
\newblock Aggregated residual transformations for deep neural networks.
\newblock In {\em Proceedings of the IEEE Conference on Computer Vision and
  Pattern Recognition}, pages 1492--1500, 2017.

\bibitem{yamashita2005application}
Takashi Yamashita, Kotaro Hirasawa, and Jinglu Hu.
\newblock Application of multi-branch neural networks to stock market
  prediction.
\newblock In {\em Proceedings. 2005 IEEE International Joint Conference on
  Neural Networks, 2005.}, volume~4, pages 2544--2548. IEEE, 2005.

\bibitem{yamashita2002multi}
Takashi Yamashita, Kotaro Hirasawa, Jinglu Hu, and Junichi Murata.
\newblock Multi-branch structure of layered neural networks.
\newblock In {\em Proceedings of the 9th International Conference on Neural
  Information Processing, 2002. ICONIP'02.}, volume~1, pages 243--247. IEEE,
  2002.

\end{thebibliography}

\end{document}